\documentclass[11pt]{article}
\usepackage[]{naacl2021}
\usepackage{times}
\usepackage{latexsym}

\usepackage{todonotes}
\usepackage{multirow}
\usepackage{graphicx}
\usepackage{enumitem}
\usepackage{anyfontsize}
\usepackage{url}
\usepackage{subcaption}
\usepackage{xr}
\usepackage{tabu}
\usepackage{bbold}
\usepackage{booktabs}
\usepackage{amsmath}
\usepackage{xcolor,pifont}
\usepackage{MnSymbol,wasysym}
\usepackage{fontawesome}
\usepackage{xspace}
\usepackage{caption}

\usepackage{microtype}
\def\aclpaperid{50} 

\newcommand{\Skip}[1]{}

\newcommand{\modelname}{EventPlus}

\newcommand{\secref}[1]{\S\ \ref{#1}}
\newcommand{\figref}[1]{Figure \ref{#1}}
\newcommand{\tbref}[1]{Table \ref{#1}}

\newcommand{\genia}[1]{GENIA}
\newcommand{\ftscibert}[1]{SciBERT-FT}

\newcommand{\dotieconcat}[2]{
  \text{\raisebox{.8ex}{$\smallfrown$}}%
}

\newcommand{\sysname}{EventPlus\xspace}

\title{\sysname: A Temporal Event Understanding Pipeline}

\author{
    Mingyu Derek Ma\textsuperscript{\rm 1}\thanks{~~Equal contribution.}~~~
    Jiao Sun\textsuperscript{\rm 2}\footnotemark[1]~~~
    Mu Yang\textsuperscript{\rm 3}~~~
    Kung-Hsiang Huang\textsuperscript{\rm 2}~~~
    \\
    {\bfseries Nuan Wen\textsuperscript{\rm 2}~~~
    Shikhar Singh\textsuperscript{\rm 2}~~~
    Rujun Han\textsuperscript{\rm 2}~~~
    Nanyun Peng\textsuperscript{\rm 1,2}}
    \\
    \textsuperscript{\rm 1} Computer Science Department, University of California, Los Angeles \\
    \textsuperscript{\rm 2} Information Sciences Institute, University of Southern California\\
    \textsuperscript{\rm 3} Texas A\&M University\\
    {\tt \{ma,violetpeng\}@cs.ucla.edu} \\
    {\tt \{jiaosun,kunghsia,nuanwen,ssingh43,rujunhan\}@usc.edu} \\ {\tt yangmu@tamu.edu}
}

\date{}

\begin{document}
\maketitle
\begin{abstract}
    We present \sysname,  a temporal event understanding pipeline that integrates various state-of-the-art event understanding components including \emph{event trigger and type detection}, \emph{event argument detection}, \emph{event duration} and \emph{temporal relation extraction}. 
Event information, especially event temporal knowledge, is a type of common sense knowledge that helps people understand how stories evolve and provides predictive hints for future events. 
\sysname{} as the first comprehensive temporal event understanding pipeline provides a convenient tool for users to quickly obtain annotations about events and their temporal information for any user-provided document.
Furthermore, we show \sysname can be easily adapted to other domains (e.g., biomedical domain). We make \sysname publicly available to facilitate event-related information extraction and downstream applications.


\end{abstract}

\section{Introduction}
\label{sec:intro}

\begin{figure}
    \centering
    \includegraphics[width=0.48\textwidth]{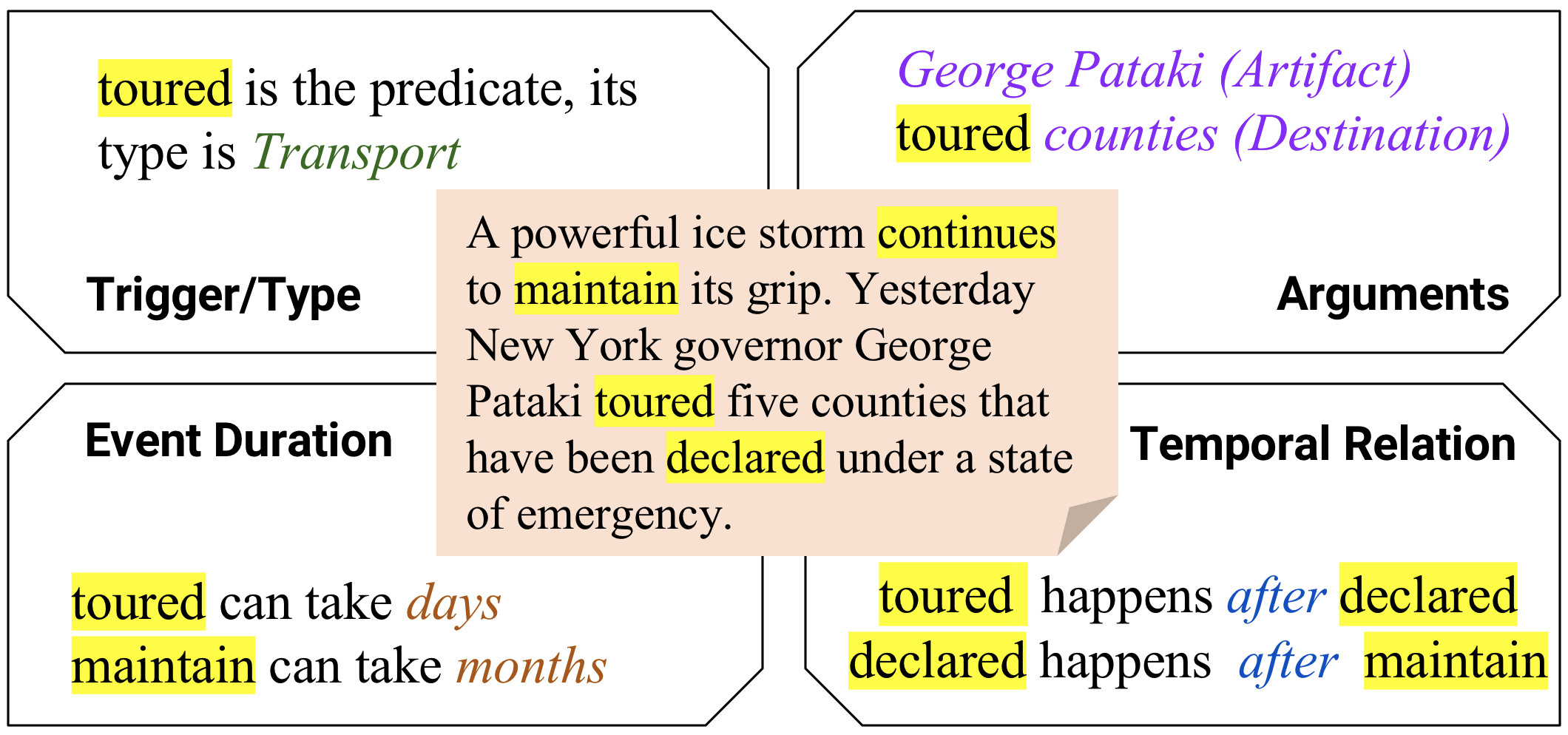}
    \caption{Event understanding components. We highlight events triggers in yellow, and mark the predicted task-related information in \emph{Italic}.}
    \label{fig:intro-example}
\end{figure}

Event understanding is intuitive for humans and important for daily decision making. For example, given the raw text shown in \figref{fig:intro-example}, a person can infer lots of information including \emph{event trigger and type}, \emph{event related arguments} (e.g., agent, patient, location), \emph{event duration} and \emph{temporal relations} between events based on the linguistic and common sense knowledge. These understandings help people comprehend the situation and prepare for future events. The event and temporal knowledge are helpful for many downstream applications including question answering~\cite{meng2017temporal,huang2019cosmos}, story generation~\cite{peng2018towards, yao2019plan, goldfarb2019plan, goldfarb2020content}, and forecasting~\cite{wang2017integrating, granroth2016happens, li2018constructing}. 

Despite the importance, there are relatively few tools available for users to conduct text-based temporal event understanding. Researchers have been building natural language processing (NLP) analysis tools for ``core NLP'' tasks~\cite{gardner2018allennlp, manning2014stanford, khashabi2018cogcompnlp}. However, systems that target at semantic understanding of events and their temporal information are still under-explored. There are individual works for event extraction, temporal relation detection and event duration detection, but they are separately developed and thus cannot provide comprehensive and coherent temporal event knowledge.

\begin{figure*}[t]
    \centering
    
    \includegraphics[width=0.9\textwidth]{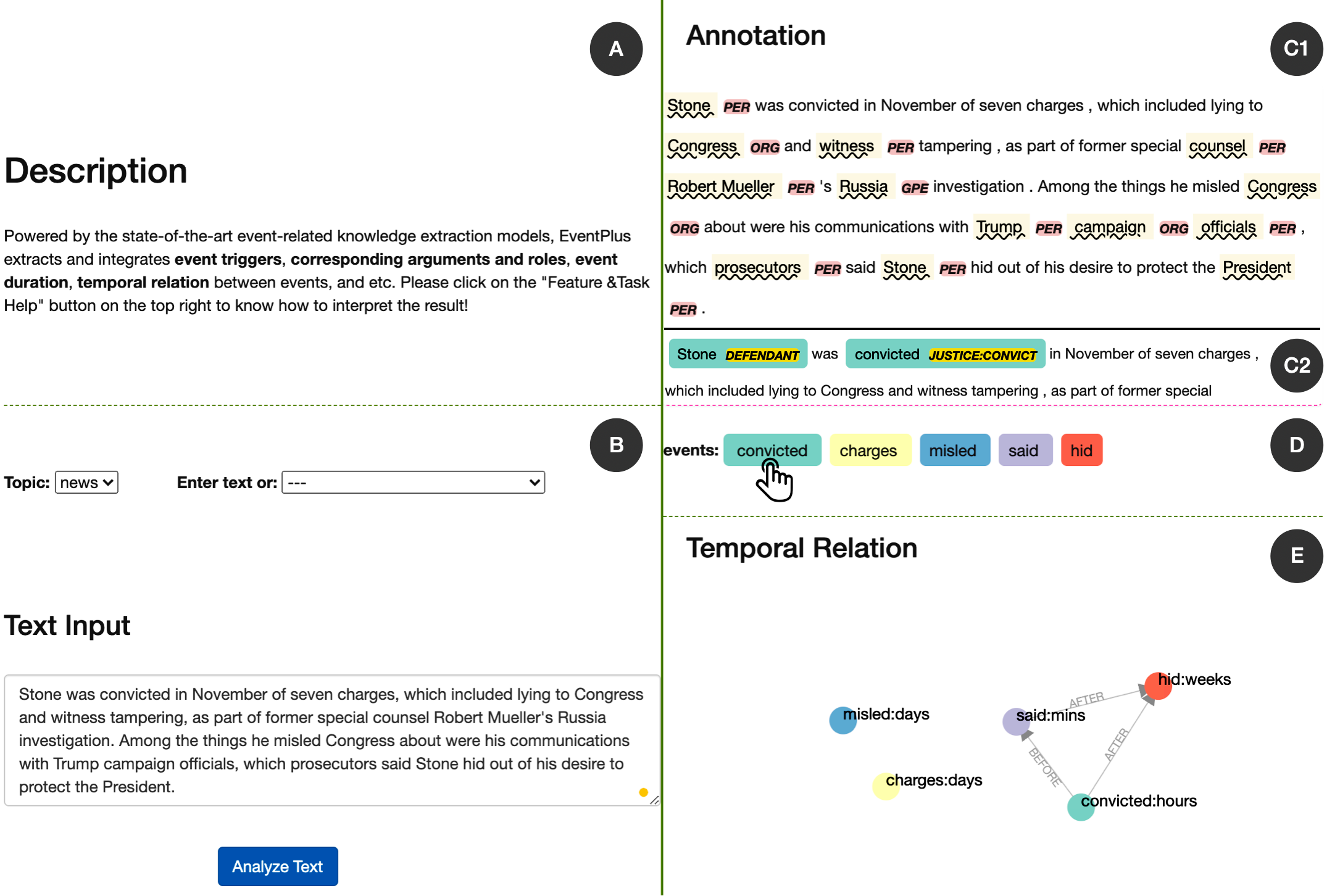}
    \caption{The interface of \sysname. Users can either choose examples or freely input text which matches with their selected topic in $B$. $C$ shows the Name Entity Recognition (NER) results, which serve as argument candidates for events. When clicking on an event trigger in $D$, we show the selected event trigger and its corresponding \emph{arguments} in $C2$. We show \emph{temporal-related information} of all events in $E$, where nodes represent event triggers and edges represent their relations; we further indicate the \emph{event duration} as labels of nodes.}
    \label{fig:screenshot}
\end{figure*}

We present \modelname, the \textit{first} pipeline system integrating several high-performance temporal event information extraction models for comprehensive temporal event understanding. Specifically, \modelname\ contains event extraction (both on defined ontology and for novel event triggers), event temporal relation prediction, event duration detection and event-related arguments and named entity recognition, as shown in \figref{fig:screenshot}.\footnote{The system is publicly accessible at \url{https://kairos-event.isi.edu}.
We also provide an introductory video at \url{https://pluslabnlp.github.io/eventplus}.}

\sysname is designed with multi-domain support in mind. Particularly, we present an initial effort to adapt \sysname to the biomedical domain. We summarize the contributions as follows:

\begin{itemize}[leftmargin=*]
\itemsep-0.2em 
    \item We present the first event pipeline system with comprehensive event understanding capabilities to extract \emph{event triggers and argument}, \emph{temporal relations} among events and \emph{event duration} to provide an event-centric natural language understanding (NLU) tool to facilitate downstream applications.
    \item Each component in \sysname has comparable performance to the state-of-the-art, which assures the quality and efficacy of our system for temporal event reasoning.
\end{itemize}


\section{Component}
\label{sec:component}
\begin{figure*}
    \centering
    \includegraphics[width=\textwidth]{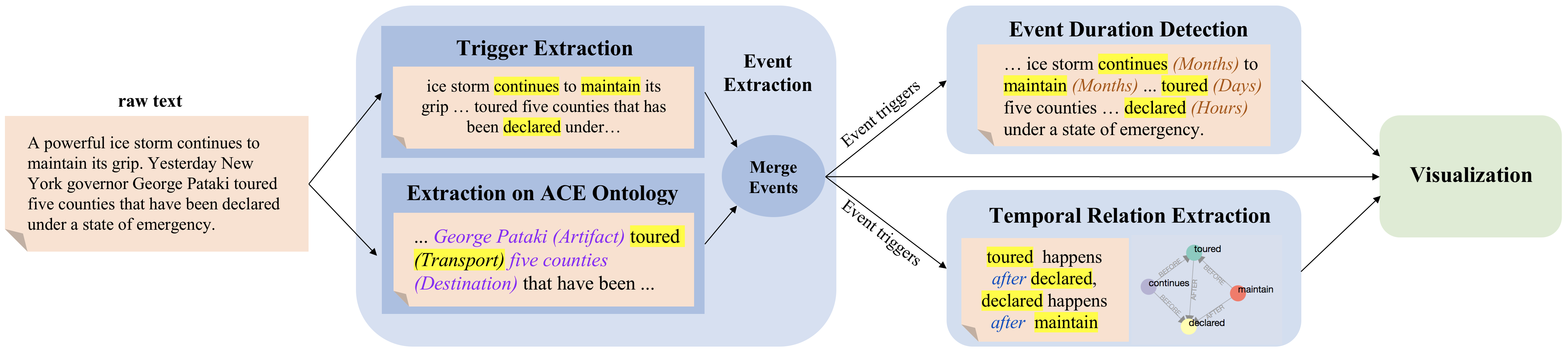}
    \caption{Overall system design of \modelname. The raw text is first fed into two event extraction components, and then we pass the event triggers of the merged event list to event duration detection and temporal relation extraction models. Finally outputs from all models are combined for visualization.}
    \label{fig:system-design}
\end{figure*}

In this section, we introduce each component in our system, as shown in \figref{fig:system-design}. We use a multi-task learning model for event trigger and temporal relation extraction (\secref{sec:event-trigger-extraction}). The model introduced in \secref{subsec:ee_ace} extracts semantic-rich events following the ACE ontology, and the model introduced in \secref{subsec:event-duration} predicts the event duration. Note that our system handles two types of event representations: one represents an event as the trigger word \cite{pustejovsky2003timeml} (as the event extraction model in \secref{sec:event-trigger-extraction}), the other represents event as a complex structure including trigger, type and arguments \cite{ahn2006stages} (as the event extraction model in \secref{subsec:ee_ace}). The corpus following the former definition usually has a broader coverage while the latter can provide richer information. Therefore, we develop models to combine the benefits of both worlds. We also introduce a speculated and negated events handling component in \secref{subsec:negation-detection} to further identify whether an event happens or not.

\subsection{Multi-task Learning of Event Trigger and Temporal Relation Extraction}
\label{sec:event-trigger-extraction}

The event trigger extraction component takes the input of raw text and outputs single-token event triggers. The input to the temporal relation extraction model is raw text and a list of detected event triggers. The model will predict temporal relationships between each pair of events. In previous literature \cite{han-etal-2019-joint}, multi-task learning of these two tasks can significantly improve performance on both tasks following the intuition that event relation signals can be helpful to distinguish event triggers and non-event tokens.

The model feeds BERT embedding \cite{devlin2019bert} of the input text to a shared BiLSTM layer for encoding task-specific contextual information. The output of the BiLSTM is passed to an event scoring function and a relation scoring function which are MLP classifiers to calculate the probability of being an event (for event extraction) or a probability distribution over all possible relations (for temporal relation extraction). We train the multi-task model on MATRES \citep{NingWuRo18} containing temporal relations \textsc{before}, \textsc{after}, \textsc{simultaneous} and \textsc{vague}. 
Though the model performs both tasks during training, it can be separately used for each individual task during inference.

\subsection{Event Extraction on ACE Ontology}
\label{subsec:ee_ace}
    
    
Although event triggers present the occurrence of events, they are not sufficient to demonstrate the semantic-rich information of events. ACE 2005\footnote{\url{https://www.ldc.upenn.edu/collaborations/past-projects/ace}} corpus defines an event ontology that represents an event as a structure with triggers and corresponding event arguments (participants) with specific roles \cite{doddington4r}.\footnote{The ACE program provides annotated data for five kinds of extraction targets: entities, times, values, relations and events. We only focus on events and entities data in this paper.} Our system is trained with ACE 2005 corpus, thus it is capable of extracting events with the complex structure. ACE focuses on events of a particular set of types including \textsc{life}, \textsc{movement}, \textsc{transaction}, \textsc{business}, \textsc{conflict}, \textsc{contact},  \textsc{personnel} and \textsc{justice}, where each type has corresponding sub-types. Following prior works \cite{wadden-etal-2019-entity, lin-etal-2020-joint}, we keep 7 entity types (person, organization, location, geo-political entity, facility, vehicle, weapon), 33 event sub-types, and 22 argument roles that are associated with sub-types.

Similar to \newcite{han-etal-2019-joint}, we build our event extraction component for ACE ontology upon a multi-task learning framework that consists of trigger detection, argument role detection and entity detection. These tasks share the same BERT encoder, which is fine-tuned during training. The entity detector predicts the argument candidates for all events in an input sentence. The trigger detector labels the input sequence with the event sub-types at the token level. The argument role detector finds the argument sequence\footnote{Argument sequences are presented using BIO encoding.} for each detected trigger via attention mechanism. For example, for the sentence in \figref{fig:intro-example}, its target trigger sequence has \textsc{Movement:Transport} label at the position of ``toured" token, and its argument sequence for this \textsc{Movement:Transport} event has \textsc{B-artifact, I-artifact} labels at the position of ``George Pataki” and \textsc{B-destination} label at the position of ``counties" respectively. The entire multi-task learning framework is jointly trained.

During inference, our system detects arguments solely based on triggers. To make our system better leverage information from argument candidates, we developed the following constraints during decoding based on the predicted entities (argument candidates) and other specific definitions in ACE:
\vspace{-2mm}
\begin{itemize}[leftmargin=*]
\itemsep-0.2em 
    \item Entity-Argument constraint. The argument role label for a token can take one of the 22 argument roles if and only if the token at this position belongs to a predicted entity.
    \item Entity-Trigger constraint. The trigger label for a token can take one of the 33 event sub-types if and only if the token at this position does not belong to a predicted entity.
    \item Valid Trigger-Argument constraint. Based on the definitions in ACE05, each event sub-type takes certain types of argument roles. We enforce that given the predicted trigger label, the argument roles in this sequence can only take those that are valid for this trigger. 
\end{itemize}
\vspace{-2mm}
To account for these constraints, we set the probability of all invalid configurations to be 0 during decoding.

\subsection{Event Duration Detection}
\label{subsec:event-duration}
        This component classifies event triggers into duration categories. While many datasets have covered time expressions which are explicit timestamps for events \cite{pustejovsky2003timebank, cassidy2014annotation, TACL1218, bethard-etal-2017-semeval}, they do not target categorical event duration. To supplement this, \newcite{vashishtha-etal-2019-fine} introduces the UDS-T dataset, where they provide 11 duration categories which we adopt for our event pipeline:  \textsc{instant},  \textsc{seconds},  \textsc{minutes},  \textsc{hours},  \textsc{days},  \textsc{weeks},  \textsc{months},  \textsc{years},  \textsc{decades},  \textsc{centuries} and \textsc{forever}. \citet{Pan2006LearningED} also present a news domain duration annotation dataset containing 58 articles developed from TimeBank corpus (we refer as Typical-Duration in the following), it provides 7 duration categories (a subset of the 11 categories in UDS-T from \textsc{seconds} to \textsc{years}).
        
        We developed two models for the event duration detection task. For a sentence, along with predicate root and span, the models perform duration classification. In the first method, we fine-tune a BERT language model \cite{devlin2019bert} on single sentences and take hidden states of event tokens from the output of the last layer, then feed into a multi-layer perceptron for classification.
        
        The second model is adapted from the UDS-T baseline model, which is trained under the multi-task objectives of duration and temporal relation extraction. The model computes ELMo embeddings \cite{peters2018deep} followed by attention layers to compute the attended representation of the predicate given sentence. The final MLP layers extract the duration category. Even though this model can detect temporal relations, it underperforms the model we described in \secref{sec:event-trigger-extraction}, so we exclude the temporal relation during inference.
        
        

\subsection{Negation and Speculation Cue Detection and Scope Resolution}
\label{subsec:negation-detection}

The event extraction components described above are designed to extract all possible events, but we identify events that are indicated by speculation (e.g., \emph{would}) or negation (e.g., \emph{not}) keywords~\cite{konstantinova2012review}. Since those events do not happen, we mark them with special labels. For example, in the sentence ``The United States is \textit{not} considering sending troops to Mozambique'', we identify ``send'' will not happen.

We adapt the BERT-based negation and speculation cue detection model and the scope resolution model introduced by \citet{Khandelwal2020NegBERTAT}. To fine-tune these models, we use the SFU Review dataset with negation and speculation annotations \cite{taboada2006methods, taboada2004analyzing, konstantinova2012review}, and we feed ground truth negation and speculation cues as input for the scope resolution model. We evaluate the two models on a separate testing set of the SFU Review dataset. The cue detection model yields a 0.92 F1 score, and the scope resolution model yields a 0.88 F1 score for token-level prediction, given ground truth cues as input. In \sysname, we input cues detected by the cue detection model to the scope resolution model.

\section{System}
\label{sec:system}

We design a pipeline system to enable the interaction among components with state-of-the-art performance introduced in \secref{sec:component} and provide a comprehensive output for events and visualize the results. \figref{fig:system-design} shows the overall system design.

\subsection{Pipeline Design}
\paragraph{Event Extraction} \sysname takes in raw text and feeds the tokenized text to two event extraction modules trained on ACE ontology-based datasets and free-formatted event triggers. The ACE ontology extraction modules will produce the output of event triggers (``toured'' is a trigger), event type (it is a \textsc{Movement:Transport} event), argument and its role (the \textsc{artifact} is ``George Pataki'' and \textsc{destination} is ``counties'') and NER result (``New York'' and ``counties'' are \textsc{geo-political entity} and ``governer'' and ``George Pataki'' are \textsc{person}). The trigger-only extraction model will produce all event triggers (``continues'', ``maintain'' and ``declared'' are also event triggers but we do not have arguments predicted for them). Then trigger-only events will be merged to ACE-style events list and create a combined event list from the two models. For each extracted event, if it is in the negation or speculation scope predicted by the cue detection and scope resolution component, then we add a ``speculation or negation'' argument to that event.
\paragraph{Duration Detection and Temporal Relation Extraction} The combined events list will be passed to the event duration detection model to detect duration for each of the extracted events (``tours'' will take \textsc{days} etc.) and passed to temporal relation extraction component to detect temporal relations among each pair of events (``toured'' is after ``declared'' etc.). Note that duration and temporal relation extraction are based on the context sentence besides the event triggers themselves and they are designed to consider contextualized information contained in sentences. Therefore ``take (a break)'' can take \textsc{minutes} in the scenario of ``Dr.\ Porter is now taking a break and will be able to see you soon'' but take \textsc{days} in the context of ``Dr.\ Porter is now taking a Christmas break'' \cite{ning2019understanding}. 
\paragraph{Visualization} To keep the resulted temporal graph clear, 
we remove predicted \textsc{vague} relations since that indicates the model cannot confidently predict temporal relations for those event pairs. Finally, all model outputs are gathered and pass to the front-end for visualization.

\subsection{Interface Design}
\figref{fig:screenshot} shows the interface design of \sysname.\footnote{We have a walk-through instruction available to help first-time end users get familiar with EventPlus. Please see our video for more information.} We display the NER result with wavy underlines and highlight event triggers and corresponding arguments with the same color upon clicks. Besides, we represent the temporal relations among events in a directed graph using d3 \footnote{\url{https://d3js.org/}} if there are any, where we also indicate each event's duration in the label for each event node.

\section{Evaluation}
\label{sec:eval}

Each capability in the pipeline has its own input and output protocol, and they require various datasets to learn implicit knowledge independently. In this section, we describe the performance for each capability on corresponding labeled datasets.

\subsection{Event Trigger Extraction}
We report the evaluation about event triggers extraction component on TB-Dense \cite{cassidy2014annotation} and MATRES \cite{NingWuRo18}, two event extraction datasets in the news domain \cite{han-etal-2019-joint}. We show the result in \tbref{table:eval_event_extration_triggers}. 
Comparing the performance on TB-Dense with CAEVO \cite{chambers-etal-2014-dense}, DEER \cite{han2020deer} and MATRES performance with \citet{ning-etal-2018-cogcomptime}, the model we use achieves best F1 scores and yields the state-of-the-art performance.
\begin{table}[h]
    \centering
     \renewcommand{\tabcolsep}{1.5mm}
    \begin{tabular}{l|l|c}
        Corpus & Model & F1 \\ \hline
        \multirow{3}{*}{TB-Dense} & \citet{chambers-etal-2014-dense} &  87.4 \\
        & \citet{han2020deer} & 90.3 \\
        & Ours & \textbf{90.8}  \\ \hline 
        \multirow{2}{*}{MATRES} &  \citet{ning-etal-2018-cogcomptime} & 85.2  \\
        & Ours & \textbf{87.8} 
    \end{tabular}
    \caption{Evaluation for event trigger extraction}
    \label{table:eval_event_extration_triggers}
\end{table}


\subsection{Event Extraction on ACE Ontology}
    
We evaluate our event extraction component on the test set of ACE 2005 dataset using the same data split as prior works \cite{lin-etal-2020-joint, wadden-etal-2019-entity}. We follow the same evaluation criteria: 
\begin{itemize}[leftmargin=*]
\itemsep-0.2em 
    \item Entity: An entity is correct if its span and type are both correct.
    \item Trigger: A trigger is correctly \textbf{identified} (Trig-I) if its span is correct. It is correctly \textbf{classified} (Trig-C) if its type is also correct.
    \item Argument: An argument is correctly \textbf{identified} (Arg-I) if its span and event type are correct. It is correctly \textbf{classified} (Arg-C) if its role is also correct. 
\end{itemize}
In Table \ref{tab:ace_result}, we compare the performance of our system with the current state-of-the-art method OneIE \cite{lin-etal-2020-joint}. Our system outperforms OneIE in terms of entity detection performance. However our trigger and argument detection performance is worse than it. We leave the improvements for triggers and arguments for future work.
\begin{table}[h]
    \centering
        \begin{tabular}{@{\ \ }l | c@{\ \ }c@{\ \ }c@{\ \ }c@{\ \ }c@{\ \ }}
            Model & Entity & Trig-I  & Trig-C  & Arg-I & Arg-C \\
            \hline
            
            OneIE & 90.2 & \textbf{78.2} & \textbf{74.7}  & \textbf{59.2} & \textbf{56.8} \\
            Ours & \textbf{91.3} & 75.8 & 72.5  & 57.7 & 55.7 \\
        \end{tabular}
    \caption{Test set performance on ACE 2005 dataset. Following prior works, we use the same evaluation criteria: *-I represent Trigger or Argument Identification. *-C represent Trigger or Argument Classification.}
    \label{tab:ace_result}
\end{table}

\subsection{Event Duration Detection}

    We evaluate the event duration detection models on Typical-Duration and newly annotated ACE-Duration dataset to reflect the performance on generic news domain for which our system is optimized. Since UDS-T dataset \cite{vashishtha-etal-2019-fine} is imbalanced and has limited samples for some duration categories, we do not use it as an evaluation benchmark but we sample 466 high IAA data points as training resources. We split Typical-Duration dataset and use 1790 samples for training, 224 for validation and 224 for testing.
    
    To create ACE-Duration, we sample 50 unique triggers with related sentences from the ACE dataset, conduct manual annotation with three annotators and take the majority vote as the gold duration category. Given natural ordering among duration categories, the following metrics are employed: accuracy over 7 duration categories (Acc), coarse accuracy (Acc-c, if the prediction falls in categories whose distance to the ground truth is 1, it is counted as correct) and Spearman correlation (Corr).

                
    
    \begin{table}[h]
        \centering
        \small
            \resizebox{\linewidth}{!}{
            \begin{tabular}{l|c@{ }| c@{ } |c@{ } |c@{ } | c@{ } | c@{ }}
                & \multicolumn{3}{c|}{Typical-Duration} & \multicolumn{3}{c}{ACE-Duration} \\
                \cline{2-7}
                Model & Acc & Acc-c & Corr  & Acc & Acc-c & Corr   \\
                \hline

                UDS-T (U) & 0.20 & 0.54 & 0.59 & 0.38 & 0.68 & 0.62\\
                UDS-T (T)& 0.52 & 0.79 & 0.71  &  0.47 & 0.67 & 0.50 \\
                UDS-T (T+U) & 0.50 & 0.76 & 0.68 & \textbf{0.49} & 0.74 & 0.66 \\
                BERT (T)& \textbf{0.59} & \textbf{0.81} & \textbf{0.75} & 0.31 & 0.67 & 0.64 \\
                BERT (T+U)& 0.56 & \textbf{0.81} & 0.73 & 0.45 & \textbf{0.79} & \textbf{0.70} 
                
            \end{tabular}
            }
        \caption{Event duration detection experimental result. Typical-Duration results are from testing subset. Notations in the bracket of model names indicate resources for training, U: 466 UDS-T high IAA samples, T: Typical-Duration training set}
        \label{tab:dur_result}
    \end{table}

    Experimental results in \tbref{tab:dur_result} show the BERT model is better than UDS-T ELMo-based model in general and data augmentation is especially helpful to improve performance on ACE-Duration. Due to the limited size of ACE-Duration, we weight more on the Typical-Duration dataset and select BERT (T) as the best configuration. To the best of our knowledge, this is the state-of-the-art performance on the event duration detection task.
\subsection{Temporal Relation Extraction}
\label{subsec:eval_temprel}
We report temporal relation extraction performance on TB-Dense and MATRES datasets. TB-Dense consider the duration of events so the labels are \textsc{includes}, \textsc{included in}, \textsc{before}, \textsc{after}, \textsc{simultaneous} and \textsc{vague}, while MATRES uses start-point as event temporal anchor and hence its labels exclude \textsc{includes} and \textsc{included in}. In \sysname, we augment extracted events from multiple components, so we report temporal relation extraction result given golden events as relation candidates to better reflect single task performance. 
\vspace{-6mm}
\begin{table}[h]
    \centering
    \resizebox{\linewidth}{!}{
    \begin{tabular}{l|l|c}
        Corpus & Model & F1 \\ \hline
        \multirow{3}{*}{TB-Dense} &
        \citet{vashishtha-etal-2019-fine} & 56.6 \\
        & \citet{meng-rumshisky-2018-context} & 57.0 \\
        & Ours & \textbf{64.5} \\ \hline 
        \multirow{3}{*}{MATRES} &  \citet{ning-etal-2018-cogcomptime} & 65.9 \\
        & \citet{NingWuRo18} & 69.0 \\
        & Ours & \textbf{75.5}
    \end{tabular}
    }
    \caption{Experimental result for temporal relation extraction given golden event extraction result}
    \label{tab:eval_temprel}
    \vspace{-3mm}
\end{table}

\tbref{tab:eval_temprel} shows the experimental results.\footnote{The MATRES experiment result in \tbref{tab:eval_temprel} uses 183 documents for training and 20 for testing developed from the entire TempEval-3 dataset. \citet{han-etal-2019-deep} reports higher F1 score but it uses a subset of MATRES (22 documents for train, 5 for dev and 9 for test) and has different setting.} Our model in \secref{sec:event-trigger-extraction} achieves the best result on temporal relation extraction and is significantly better than \cite{vashishtha-etal-2019-fine} mentioned in \secref{subsec:event-duration}.\footnote{
The latest state-of-the-art work \cite{han2020deer} only reports end-to-end event extraction and temporal relation extraction result, pure temporal relation extraction result given ground-truth events are not provided.
We are not able to compare with it directly.}

\section{Extension to Biomedical Domain}
\label{sec:biomed}

With our flexible design, each component of \sysname can be easily extended to other domains with little modification. We explore two approaches to extend the event extraction capability (\secref{subsec:ee_ace}) to the biomedicine domain: 1) multi-domain training (MDT) with \genia{} \cite{kim-etal-2009-overview}, a dataset containing biomolecular interaction events from scientific literature, with shared token embeddings, which enables the model to predict on both news and biomedical text;
 2) replace the current component with an in-domain event extraction component \textbf{\ftscibert{}} \cite{huang-etal-2020-biomedical} which is a biomedical event extraction system based on fine-tuned SciBERT \cite{beltagy-etal-2019-scibert}.

\begin{figure}[h]
\centering
    \includegraphics[width=0.46\textwidth]{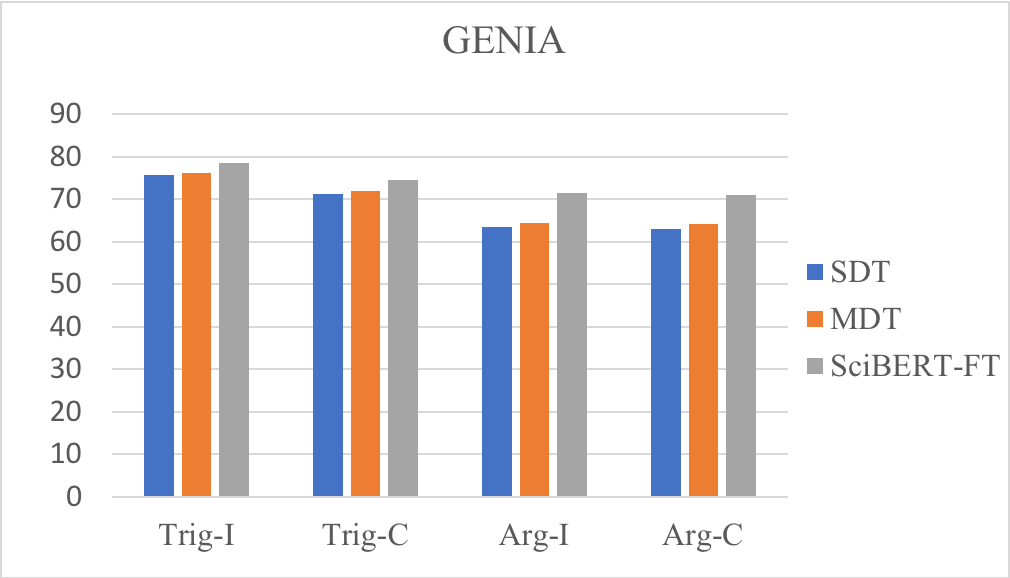}
    \caption{Performance comparison of single-domain training (SDT), multi-domain training (MDT) and \ftscibert{} on the Dev set of \genia{}}
    \label{fig:mdt_1}
\end{figure}

While MDT on ACE and GENIA datasets from different domains improves the performance on \genia{}, it is still lower than \textbf{\ftscibert{}} (\figref{fig:mdt_1}). 
Therefore, we decide to pursue the second extension approach to incorporate \textbf{\ftscibert{}} and extend \sysname to the biomedical domain.


\section{Related Works}
\label{sec:relatedworks}

Existing NLP toolkits \cite{manning2014stanford, khashabi2018cogcompnlp} provide an interface for a set of useful models. Some tools integrate several models in a pipeline fashion \cite{peng-etal-2015-concrete, noji-miyao-2016-jigg}. The majority of these systems focus on token-level tasks like tokenization, lemmatization, part-of-speech tagging, or sentence-level tasks like syntactic parsing, semantic role labeling etc.
There are only a few systems that can provide capabilities of event extraction and temporal information detection \cite{tao2013eventcube, ning2019understanding}. 

For event extraction, some systems only provide results within a certain defined ontology such as AIDA
\cite{li-etal-2019-multilingual}, there are also some works utilizing data from multiple modalities \cite{li-etal-2020-gaia, li-etal-2020-cross}. Some works could handle novel events \cite{xiang2019survey, ahmad2021gate, han2020knowledge, Huang2020EfficientEL}, but they are either restricted to a certain domain \cite{yang-etal-2018-dcfee} or lack of performance superiority because of their lexico-syntactic rule-based algorithm \cite{valenzuela-escarcega-etal-2015-domain}. For temporal information detection, \citet{ning-etal-2019-improved} proposes a neural-based temporal relation extraction system with knowledge injection. Most related to our work, \citet{ning-etal-2018-cogcomptime} demonstrates a temporal understanding system to extract time expression and implicit temporal relations among detected events, but this system cannot provide event-related arguments, entities and event duration information.

These previous works either are not capable of event understanding or just focus on one perspective of event-related features. There is no existing system that incorporates a comprehensive set of event-centric features, including event extraction and related arguments and entities, temporal relations, and event duration.


\section{Conclusion and Future Work}
\label{sec:conclusion}
We represent \sysname, a pipeline system that takes raw texts as inputs and produces a set of temporal event understanding annotations, including \emph{event trigger and type}, \emph{event arguments}, \emph{event duration} and \emph{temporal relations}. To the best of our knowledge, \sysname is the first available system that provides such a comprehensive set of temporal event knowledge extraction capabilities with state-of-the-art components integrated. 
We believe \sysname will provide insights for understanding narratives and facilitating downstream tasks.

In the future, we plan to further improve \sysname\ by tightly integrating event duration prediction and temporal relation extraction modules. We also plan to improve the performance for triggers and arguments detection under the ACE ontology and develop joint training models to optimize all event-related features in an end-to-end fashion.

\section*{Acknowledgments}
Many thanks to Yu Hou for the quality assessment annotations, to Fred Morstatter and Ninareh Mehrabi for feedback on the negation and speculation event handling, and to the anonymous reviewers for their feedback.
This material is based on research supported by DARPA under agreement number FA8750-19-2-0500. The U.S. Government is authorized to reproduce and distribute reprints for Governmental purposes notwithstanding any copyright notation thereon. The views and conclusions contained herein are those of the authors and should not be interpreted as necessarily representing the official policies or endorsements, either expressed or implied, of DARPA or the U.S. Government. 


\bibliographystyle{acl_natbib}
\bibliography{emnlp2020}




\end{document}